\documentclass[letterpaper]{article} 
\usepackage{aaai25}  
\usepackage{times}  
\usepackage{helvet}  
\usepackage{courier}  
\usepackage[hyphens]{url}  
\usepackage{graphicx} 
\usepackage{natbib}  
\usepackage{caption} 
\usepackage{algorithm}
\usepackage{algorithmic}
\usepackage{amsmath} 
\usepackage{amssymb}

\usepackage{cases}
\usepackage{bm}
\usepackage{algorithm}
\usepackage{algorithmic}
\usepackage{cases}
\usepackage{multirow}
\usepackage{color} 
\usepackage{enumerate}
\usepackage{etoolbox}
\usepackage{mathrsfs}
\newcommand{\MyMapTemplatePrefix}[4]{\expandafter#1\csname#3#4\endcsname{#2{#4}}}
\newcommand{\MyMapTemplatePrefixNew}[5]{\expandafter#1\csname#4#5\endcsname{#2{#3{#5}}}}
\forcsvlist{\MyMapTemplatePrefix {\def} {\mathbf} {}} {A,B,C,D,E,F,G,H,I,J,K,L,M,N,O,P,Q,R,S,T,U,V,W,X,Y,Z}
\forcsvlist{\MyMapTemplatePrefix {\def} {\mathbf} {}} {a,b,c,d,e,f,g,h,i,j,k,l,m,n,o,p,q,r,s,t,u,v,w,x,y,z,1,0}
\forcsvlist{\MyMapTemplatePrefix {\def} {\widetilde} {wt}} {A,B,C,D,E,F,G,H,I,J,K,L,M,N,O,P,Q,R,S,T,U,V,W,X,Y,Z}
\forcsvlist{\MyMapTemplatePrefix {\def} {\widetilde} {wt}} {a,b,c,d,e,f,g,h,i,j,k,l,m,n,o,p,q,r,s,t,u,v,w,x,y,z} 
\forcsvlist{\MyMapTemplatePrefixNew {\def} {\widetilde}{\mathbf} {tb}} {A,B,C,D,E,F,G,H,I,J,K,L,M,N,O,P,Q,R,S,T,U,V,W,X,Y,Z}
\forcsvlist{\MyMapTemplatePrefixNew {\def} {\widetilde}{\mathbf} {tb}} {a,b,c,d,e,f,g,h,i,j,k,l,m,n,o,p,q,r,s,t,u,v,w,x,y,z}
\forcsvlist{\MyMapTemplatePrefix {\def} {\widehat} {wh}} {A,B,C,D,E,F,G,H,I,J,K,L,M,N,O,P,Q,R,S,T,U,V,W,X,Y,Z}
\forcsvlist{\MyMapTemplatePrefix {\def} {\widehat} {wh}} {a,b,c,d,e,f,g,h,i,j,k,l,m,n,o,p,q,r,s,t,u,v,w,x,y,z}
\forcsvlist{\MyMapTemplatePrefixNew {\def} {\widehat}{\mathbf} {hb}} {A,B,C,D,E,F,G,H,I,J,K,L,M,N,O,P,Q,R,S,T,U,V,W,X,Y,Z}
\forcsvlist{\MyMapTemplatePrefixNew {\def} {\widehat}{\mathbf} {hb}} {a,b,c,d,e,f,g,h,i,j,k,l,m,n,o,p,q,r,s,t,u,v,w,x,y,z}
\forcsvlist{\MyMapTemplatePrefixNew {\def} {\overline}{\mathbf} {lb}} {A,B,C,D,E,F,G,H,I,J,K,L,M,N,O,P,Q,R,S,T,U,V,W,X,Y,Z}
\forcsvlist{\MyMapTemplatePrefixNew {\def} {\overline}{\mathbf} {lb}} {a,b,c,d,e,f,g,h,i,j,k,l,m,n,o,p,q,r,s,t,u,v,w,x,y,z}
\forcsvlist{\MyMapTemplatePrefix {\def} {\mathcal}{mc}} {A,B,C,D,E,F,G,H,I,J,K,L,M,N,O,P,Q,R,S,T,U,V,W,X,Y,Z}
\forcsvlist{\MyMapTemplatePrefix {\def} {\mathbb} {mb}} {A,B,C,D,E,F,G,H,I,J,K,L,M,N,O,P,Q,R,S,T,U,V,W,X,Y,Z}

 \def\eg{{e.g.}}

\usepackage{newfloat}
\usepackage{listings}
\DeclareCaptionStyle{ruled}{labelfont=normalfont,labelsep=colon,strut=off} 
\floatstyle{ruled}
\newfloat{listing}{tb}{lst}{}
\floatname{listing}{Listing}
\title{Re-Attentional Controllable Video Diffusion Editing}
\author{
    Yuanzhi~Wang\textsuperscript{\rm 1,2}, 
    Yong~Li\textsuperscript{\rm 1,3}, 
    Mengyi~Liu\textsuperscript{\rm 2}, 
    Xiaoya~Zhang\textsuperscript{\rm 1,}\thanks{Corresponding authors: Xiaoya~Zhang and Zhen~Cui}, \\
    Xin~Liu\textsuperscript{\rm 4}, 
    Zhen~Cui\textsuperscript{\rm 1,}\footnotemark[1], 
    Antoni~B.~Chan\textsuperscript{\rm 3}
}
\affiliations{
    \textsuperscript{\rm 1}School of Computer Science and Engineering, Nanjing University of Science and Technology, Nanjing, China\\
    \textsuperscript{\rm 2}Department of Content Security, Kuaishou Technology, Beijing, China\\
    \textsuperscript{\rm 3}Department of Computer Science, City University of Hong Kong, Hong Kong, China\\
    \textsuperscript{\rm 4}SeetaCloud, Nanjing, China\\

}

\usepackage{bibentry}

\begin{document}
	
	\maketitle
	
	\begin{abstract}
		Editing videos with textual guidance has garnered popularity due to its streamlined process which mandates users to solely edit the text prompt corresponding to the source video.
		Recent studies have explored and exploited large-scale text-to-image diffusion models for text-guided video editing, resulting in remarkable video editing capabilities.
		However, they may still suffer from some limitations such as mislocated objects, incorrect number of objects.
		Therefore, the controllability of video editing remains a formidable challenge.
		In this paper, we aim to challenge the above limitations by proposing a \textit{\underline{Re}-\underline{At}tentional \underline{Co}ntrollable Video Diffusion Editing (\textbf{ReAtCo})} method.
		Specially, to align the spatial placement of the target objects with the edited text prompt in a training-free manner, we propose a Re-Attentional Diffusion (RAD) to refocus the cross-attention activation responses between the edited text prompt and the target video during the denoising stage, resulting in a spatially location-aligned and semantically high-fidelity manipulated video.
		In particular, to faithfully preserve the invariant region content with less border artifacts, we propose an Invariant Region-guided Joint Sampling (IRJS) strategy to mitigate the intrinsic sampling errors w.r.t the invariant regions at each denoising timestep and constrain the generated content to be harmonized with the invariant region content.
		Experimental results verify that ReAtCo consistently improves the controllability of video diffusion editing and achieves superior video editing performance. Codes are released at \url{https://github.com/mdswyz/ReAtCo}
	\end{abstract}

	\begin{figure}[t]
		\centering{\includegraphics[width=\linewidth]{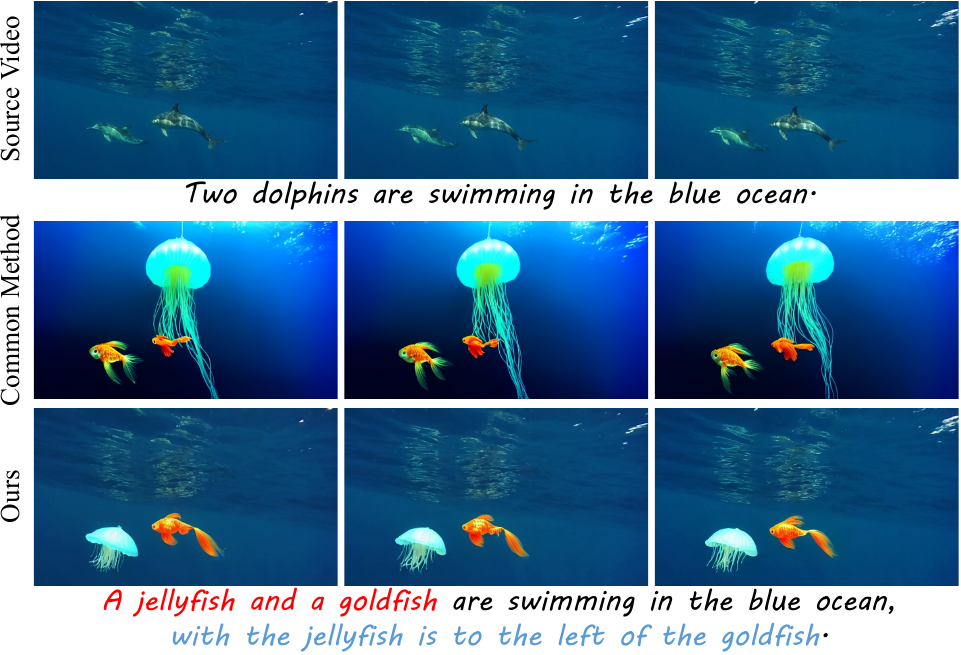}}
		\vspace{-0.5cm}
		\caption{Edited samples from the common video diffusion editing method (classic Tune-A-Video~\cite{TAV} as an example) and our proposed ReAtCo.
		}
		\label{fig:1}
		\vspace{-0.55cm}
	\end{figure}
	
	\section{Introduction}
	Text-guided video editing is a specialized facet of content creation, which can edit video content, including but not limited to manipulating objects, changing backgrounds, by manipulating the text prompt describing the source video.
	This task exemplifies the potential to augment and polish content within diverse domains, encompassing advertising design, marketing, and social media content~\cite{controlediting}.
	
	Recently, diffusion-based generative paradigm~\cite{ddpm} has shown astonishing text-to-image (T2I)~\cite{stablediffusion,imagen} and text-to-video (T2V)~\cite{Imagenvideo,Alignyourlatents} generation capabilities, which provides a great opportunity to manipulate video content via text guidance.
	To edit videos with low computational costs, some studies utilize large-scale pretrained T2I diffusion models, e.g., Stable Diffusion~\cite{stablediffusion} to develop various text-guided video editing methods~\cite{TAV,fatezero}.
	The main idea of these methods is to flatten the temporal dimensionality of the source video and diffuse the flattened video into noise, and then the inverted noise is gradually denoised to the edited videos by the T2I-based video diffusion editing model under the condition of the edited text prompt.
	Moreover, due to the inherent absence of temporal awareness in T2I diffusion models, off-the-shelf methods tend to incorporate some additional modules or mechanisms to construct a well-designed video diffusion editing model, thus preserving the temporal consistency of edited videos.
	For example, Tune-A-Video~\cite{TAV} incorporated the temporal attention modules and spatio-temporal attention modules into the T2I models for temporal coherence.
	FateZero~\cite{fatezero} proposed a fusing attention mechanism to fuse the attention maps from the diffusion and generation process to facilitate motion consistency.
	TokenFlow~\cite{TokenFlow} designed a propagation mechanism to propagate a small set of edited features across frames.
	
	Despite the great success, the controllability of editing remains a formidable challenge when performing fine-grained manipulation of multiple foreground objects.
	As shown in Fig.~\ref{fig:1}, the results from the common method show mislocated objects (i.e., the jellyfish is above the goldfish which is not aligned with \textit{``the jellyfish is to the left of the goldfish''}) and incorrect number of objects (i.e., two goldfish and a jellyfish are generated which do not match \textit{``A jellyfish and a goldfish''}).
	The essence behind this situation is the lack of spatial location awareness for the pretrained T2I models~\cite{controllableT2I,controllableT2I2}.
	A question arises: \textit{Can we improve the controllability of video editing based on off-the-shelf methods?}
	
	In this paper, we aim to challenge the above limitations by proposing a \underline{Re}-\underline{At}tentional \underline{Co}ntrollable Video Diffusion Editing (ReAtCo) method.
	To efficiently control the spatial location of the edited objects aligned with the edited text prompts in a training-free manner, a Re-Attentional Diffusion (RAD) is proposed to refocus the cross-attention activation responses between the editied prompt and video content during the denoising stage, resulting in a spatially location-aligned and semantically high-fidelity target video.
	In addition, as each denoising timestep may lead to some sampling errors~\cite{sampling-error}, the invariant region content that may exist during editing (e.g., the background in Fig.~\ref{fig:1}) is inevitably disrupted, ultimately resulting in a generated invariant region content that is far from the original ones.
	Therefore, we design an Invariant Region-guided Joint Sampling (IRJS)  strategy to mitigate the sampling errors of the invariant region by injecting the original invariant region content into the denoising process, thus maintaining the invariant region information and constraining the generated content to be harmonized with the invariant region.
	
	In contrast to prior works, our proposed ReAtCo could bring two benefits:\textbf{1)} ReAtCo can provide the ability for fine-grained manipulation of multiple foreground objects.
	As shown in Fig.~\ref{fig:1}, our ReAtCo can successfully edit \textit{``two dolphins''} into \textit{``a jellyfish and a goldfish''} while ensuring their spatial locations aligned with the target prompt (i.e., \textit{``the jellyfish is to the left of the goldfish''}).
	\textbf{2)} the invariant region content could be faithfully preserved and the generated content is harmonized with the invariant region.
	We can observe from Fig.~\ref{fig:1} that the background region (i.e., the invariant region in this case) content is consistently preserved while editing the two foreground objects.
	In summary, the contributions of this work can be concluded as:
	\begin{itemize}
		\item To improve the controllability of video editing, we propose a Re-Attentional Controllable Video Diffusion Editing (ReAtCo) method. ReAtCo can refocus the cross-attention activation responses by a well-designed RAD to control the spatial location of the edited objects aligned with the edited text prompt in a training-free manner.
		\item To keep the consistency of the invariant region with less border artifacts maximally, we design an IRJS to mitigate the sampling errors of the invariant region at each denoising timestep and to constrain the generated content to be harmonized with the invariant region.
		\item We perform extensive experiments and achieve superior or comparable results, demonstrating that our ReAtCo mitigates the limitations of existing state-of-the-arts, such as mislocated objects, incorrect number of objects.
	\end{itemize}
	
	\section{Related Works}
	
	\textbf{Text-to-image/video Generation.}
	Text-to-image (T2I) generation task aims to generate photorealistic images that semantically match given text prompts~\cite{AlignDRAW,DALL-E,stablediffusion,shen2024imagdressing}.
	The main idea of this task is to utilize the generative models~\cite{GANs,ddpm,dicmor,imder} to construct a text-conditioned generative model with various attention or Transformer mechanism~\cite{attention,li2018occlusion,zhang2020feature,li2020learning,faceformer, li2023contrastive}.
	Recently, due to powerful data generation capabilities, diffusion-based generative models have achieved great success in the T2I generation~\cite{DALLE-2,imagen,stablediffusion,MMM-RS,shen2024boosting}.
	For example, \cite{DALLE-2} proposed the DALLE-2 that uses CLIP-based~\cite{clip} feature embedding to build a T2I diffusion model with improved text-image alignments.
	\cite{stablediffusion} proposed a novel Latent Diffusion Model (LDM) paradigm that projects the original image space into the latent space of an autoencoder to improve T2I training efficiency.
	Despite the great success, text-to-video (T2V) generation is still extremely challenging due to the thousands of times harder to train compared to T2I models.
	Some researchers have attempted to challenge the T2V generation task and have proposed various methods~\cite{Imagenvideo,Pvdm,higen}.
	For instance, \cite{VDM} proposed a Video Diffusion Model that uses a space-only 3D Unet to fit video content.
	\cite{Alignyourlatents} applied the LDM to high-resolution video generation.
	
	\textbf{Controllable Text-to-image/video Generation.}
	Different from the above naive text-to-image/video methods, some studies aim to conduct controllable text-to-image/video generation by exploiting additional prior conditions~\cite{controllableT2I2,ControlVideo,shen2024imagpose,shenadvancing}.
	For example, \cite{controlnet} proposed a ControlNet that appended additional conditions, such as Canny edges, depth maps, human poses, to provide diverse image generation capabilities.
	With this work, \cite{ControlVideo} and \cite{Control-A-Video} extended the ControlNet to the video generation domain, thereby achieving controllable T2V generation.
	\cite{Reco} and \cite{refocusing} leveraged the bounding boxes to constrain the object generation. 
	\cite{Spatext} utilized the segmentation maps to control the generation regions.

	\textbf{Text-guided Video Editing.}
	The goal of text-guided video editing is to generate a new video derived from a given source video and an edited text prompt~\cite{TAV,fatezero,Stablevideo,TokenFlow}.
	Compared to earlier works such as~\cite{lna}, this technology can reduce manual labor as the users only need to edit the text prompts describing the source videos.
	Before the diffusion-based era, \cite{text2live} proposed a Text2Live to conduct text-driven video editing.
	The main idea of Text2Live is to utilize the layered neural atlas model~\cite{lna} to map source video into the image-based 2D atlas domain, thereby reducing the difficulty of video editing.
	In the era of diffusion models, \cite{Stablevideo} exploited the pretrained T2I diffusion models to edit 2D atlas images, but training the atlas models requires tremendous computational and time costs ($7\sim8$ hours for training each video).
	Another effective paradigm is to flatten the temporal dimensionality of the source video and leverage DDIM~\cite{DDIM} for the video-to-noise inversion, and then the inverted noise is gradually denoised to the edited videos by the pretrained T2I diffusion models. 
	For example, \cite{TAV} proposed a Tune-A-Video that flattens the temporal dimensionality of the source video and then edits it frame-by-frame using the T2I model to generate the target video.
	Of these, the extra temporal attention modules are injected into the T2I model to preserve the temporal consistency.
	\cite{TCVE} designed a temporal Unet to guarantee comprehensive temporal modeling.
	TokenFlow~\cite{TokenFlow} designed a cross-frame propagation mechanism to enhance the temporal smoothness.
	
	\section{Method}
	
	\subsection{Problem Description}
	
	\textbf{Problem}~~ Let $\mcV=(v_1,v_2,\cdots, v_m)$ denotes a source video that contains $m$ video frames. $\mcP$ and $\mcP'$ denote the source prompt describing $\mcV$ and the edited prompt provided by the users, respectively.
	The goal of text-guided video editing is to generate a new video
	$\mcV'$ from source video $\mcV$ under the condition of the edited prompt $\mcP'$. We illustrate an example:
	\begin{itemize}
		\item \textbf{Source:} an initial video with a prompt \textit{``Two dolphins are swimming in the blue ocean.''}
		
		\item \textbf{Target 1:} output a video to change \textit{``Two dolphins''} as \textit{``Two goldfishes''}. 
	\end{itemize}
	Recent state-of-the-art methods can excellently achieve the goal by  modifying the prompt based on the pretrained text-to-image (T2I) diffusion models, such as \textit{``Two goldfishes are swimming in the blue ocean.''} for Target 1. 
	However, the fine-grained controllability of video editing remains a formidable challenge, \eg, to simply continue the above example (a failure for most existing methods):
	\begin{itemize}
		\item \textbf{Target 2:} output a video to fine-grained manipulate \textit{``Two dolphins''} by editing \textit{``the left dolphin as a jellyfish''} and \textit{``the right dolphin as a goldfish''}.
	\end{itemize}
	The reason behind this failure is that the employed base models (i.e., the pretrained T2I models) 
	are typically trained on simple text descriptions, not including fine-grained spatial location descriptions between different objects~\cite{controllableT2I,controllableT2I2}. In other words, these methods often lack spatial location awareness for controllable video editing. 
	A question arises: \textit{Can we improve the fine-grained controllability of video editing with training-free mode?} It is not necessary to rebuild a new training dataset with information-enriched long text descriptions and retrain a new model due to high resource requirements.
	
	\textbf{Idea}~~The edited video could be partitioned into two parts: changed parts and the remaining unchanged part (\eg, background, which we denote as the invariant region). For these changed parts, the users more focus on those objects of interest, which could be decided by the input prompts $\mcP'$ and $\mcP$. Suppose $n$ objects need to be manipulated, denoted $\{O_i|_{i=1}^n\}$, the remaining part except objects is denoted $O^-$. To bridge the latent semantic information from new prompt $\mcP'$ to the video as well as keep spatial location awareness, we use text-video cross-attention maps (between text and denoised videos) to associate the objects of interest, denoted $\mcA_{O_i}(\mcV(t), \mcP')$ for object $O_i$, where $\mcV(t)$ is a noisy video at the $t$-th sampling step of the denoising process. For the unchanged part such as the background region, we expect to perform a diffusion-identical transformation $\mcD_\text{I}$ to prevent the disruption of the unchanged region. Formally, our video sampling process ($t$ to $t\!\!-\!\!1$ timestep) is defined as:
	\begin{small}
		\begin{align}\label{eq-main-idea}
			\mcV(\!t\!\!-\!\!1\!)\!\leftarrow\! F(\mcD_\text{R}(\mcV(t), \{\mcA_{O_i}(\mcV(t), \!\mcP')|_{i=1}^n\},\!\mcP'),\mcD_\text{I}(\mcV_{O^-}(t), \!\mcP')\!),  
		\end{align}
	\end{small}
	where $\mcD_\text{R}$ is the diffusion editor w.r.t the changeable objects, $F$ is an integration operation, and  $\mcV_{O^-}(t)$ is the unchanged part of $\mcV(t)$. Accordingly, there are two questions that need to be solved:
	\begin{itemize}
		\item[-] Spatial-aware diffusion editor $\mcD_\text{R}$: the spatial alignment problem between object prompts and intermediate sampled video in a training-free manner. We propose a \textbf{Re-Attentional Diffusion (RAD)}.
		\item[-] Diffusion-identical transformation $\mcD_\text{I}$: recovery unchanged region with less border artifacts when integrating with new-generated object regions. We propose an \textbf{Invariant Region-guided Joint Sampling (IRJS)}.
	\end{itemize}

	\begin{figure*}[t]
		\centering{\includegraphics[width=\linewidth]{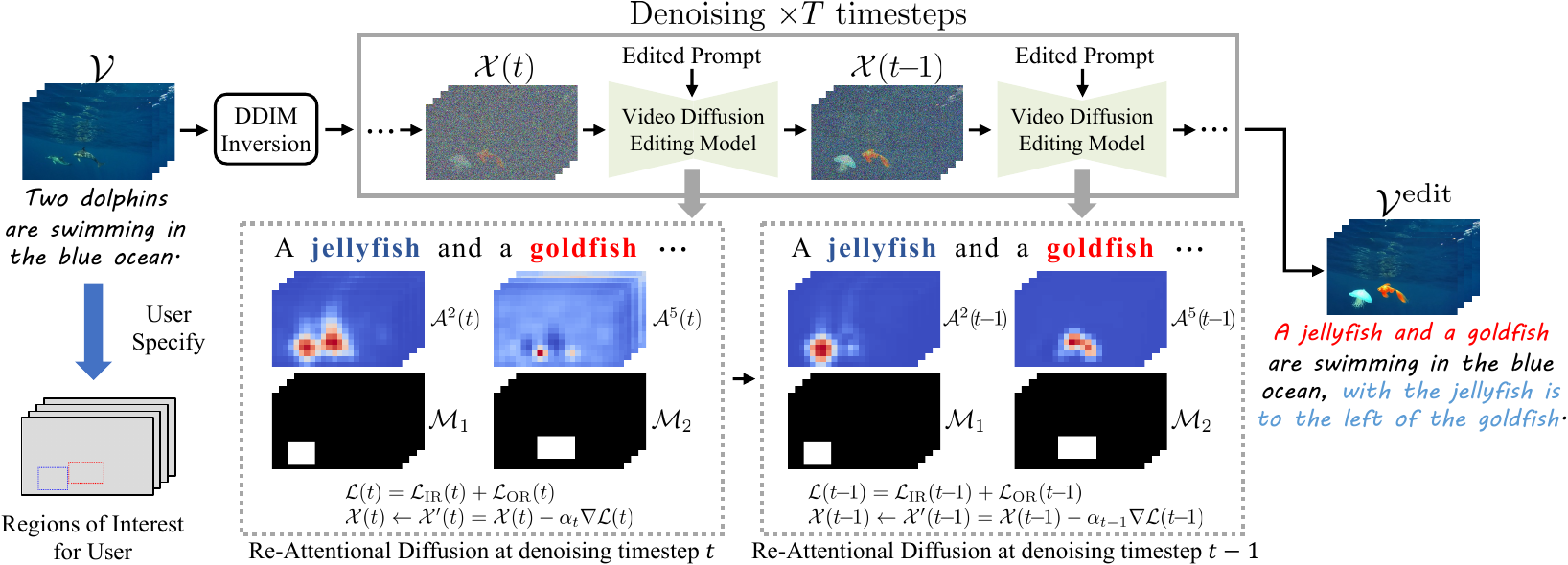}}
		\vspace{-0.5cm}
		\caption{The framework of our proposed ReAtCo. Given a source video $\mcV$, ReAtCo first utilizes DDIM Inversion for video-to-noise inversion, and then the inverted noise is gradually denoised to an edited video $\mcV^{\text{edit}}$ by a video diffusion editing model. During the denoising stage, ReAtCo injects the proposed Re-Attentional Diffusion (RAD) and the user-specified regions of interest (i.e., the regions of two dolphins $\mcM_1$, $\mcM_2$) into video diffusion editing model to refocus the cross-attention maps (e.g., $\mcA^2(t)$ and $\mcA^5(t)$ for word index $2$ and $5$ at timestep $t$) between words of interest (``jellyfish'' and ``goldfish'') and noisy video (e.g., $\mcX(t)$ at timestep $t$), thereby controlling the spatial location of the edited objects. In addition to the above, we design an Invariant Region-guided Joint Sampling (IRJS) to prevent the disruption of the invariant region with less border artifacts.
		}
		\vspace{-0.5cm}
		\label{fig:2}
	\end{figure*}
	
	\subsection{Overview Framework}
	The overview framework of ReAtCo is illustrated in Fig.~\ref{fig:2}.
	Given a source video, we first utilize DDIM Inversion~\cite{DDIM} for the video-to-noise inversion.
	Then, the inverted noise is  gradually denoised to the edited video by an off-the-shelf
	video diffusion editing model.
	In practice, we use the classic Tune-A-Video~\cite{TAV} as the video diffusion editing model to conduct experiments.
	To achieve controllable video editing, the user needs to specify the region of interest according to their edited text prompt, e.g., the regions of two dolphins in the case of Fig.~\ref{fig:2}.
	Subsequently, the region of interest can be transformed into a set of binary masks, which are injected into the denoising stage to refocus the cross-attention activation responses by our proposed RAD, resulting in a spatially location-aligned and semantically high-fidelity edited video.
	In addition, to prevent the disruption of the invariant region with less border artifacts, we propose an IRJS strategy that mitigates the sampling errors of the invariant region to maintain the original invariant content and allows the generated content to be harmonized with the invariant region.
	
	\subsection{Re-Attentional Diffusion}\label{Re-Attentional}

	Reviewing the mainstream video diffusion editing models~\cite{TAV,fatezero,Stablevideo,TCVE}, where the interaction between the textual semantic space and the pixel space occurs in the cross-attention layers of the pretrained T2I model such as Stable Diffusion~\cite{stablediffusion}.
	That means that each video frame is computed the cross-attention maps with the text embedding, thereby bridging the relationship between text and video.
	Reviewing the computation of cross-attention maps, taking the $i$-th video frame as an example, and assume that we obtain the noisy video frame feature $\mathbf{X}_{i}(t)$ at denoising timestep $t$.
	$\mathbf{X}_{i}(t)$ is multiplied by the learnable parameter $\mathbf{W}_Q$ to obtain Query $\mathbf{Q}_{i}(t)=\mathbf{W}_Q\mathbf{X}_{i}(t)\in \mathbb{R}^{H\times W\times C}$, where $H$, $W$, and $C$ indicate the height, width, and the channel dimensionality.
	The input word embedding $\mathbf{E}$ is multiplied by the learnable parameter $\mathbf{W}_K$ to generate Key $\mathbf{K}=\mathbf{W}_K\mathbf{E}\in \mathbb{R}^{L\times C}$, where $L$ is the number of text tokens.
	With $\mathbf{Q}_{i}(t)$ and $\mathbf{K}$, the cross-attention maps $\mathbf{A}_i(t)$ of the $i$-th frame at denoising timestep $t$ can be computed as:
	\begin{equation}
		\mathbf{A}_i(t)=\text{Softmax}(\mathbf{Q}_i(t)\mathbf{K^\top}/\sqrt{d})\in\mathbb{R}^{L\times H\times W}.
	\end{equation}
	From the above, $\mathbf{A}_i(t)$ is a tensor with the size of $L\times H\times W$, which means that each word is associated with a $H\times W$ pixel space cross-attention map, the values inside represent the relevance of the word to the pixel space.
	At a high level, the high response region in the cross-attention map associated with each word is equivalent to the region of generating word concept in the video frames, i.e., the higher the response, the more the word concept is being attended to in that region, and the content generated in that region is more aligned with word concept.
	
	Inspired by the above phenomenon and facts, therefore, by modifying the pixel space cross-attention map corresponding to the word of interest in $\mathbf{A}_i(t)$, we could constrain the pixel region in which the word concept is generated.
	Taking Fig.~\ref{fig:2} as an example, the user can first specify the two regions for the left and right dolphins from the source video (specify manually or automatically using the object detector), and then two regions can be transformed into two sets of binary masks $\mcM_1=\{\mathbf{M}_1^1,\mathbf{M}_2^1,\cdots\,\mathbf{M}_m^1\}$ and $\mcM_2=\{\mathbf{M}_1^2,\mathbf{M}_2^2,\cdots\,\mathbf{M}_m^2\}$.
	In this case, the ultimate goal is to edit the content of $\mcM_1$ to a jellyfish and the content of $\mcM_2$ to a goldfish.
	Thus, the words of interest are \textit{``jellyfish''} and \textit{``goldfish''} (the indexes of words are $\mcI=\{2,5\}$), and we can modify the $2$-nd and $5$-th cross-attention maps along the $L$ dimensionality of $\mathbf{A}_i(t)$ to maximize the attention response in the $\mathbf{M}_i^1$ and $\mathbf{M}_i^2$ regions, respectively. 
	Once the cross-attention maps of all video frames are carefully modified, we can obtain a spatially location-aligned and semantically high-fidelity target video.
	For modifying cross-attention maps, a simple way is to modify all responses inside the object regions to $1$ and outside the object regions to $0$, but such a straightforward way may collapse the denoising process, potentially leading to a collapse of video fidelity.
	
	Therefore, we propose a Re-Attentional Diffusion (RAD) that contains an inner-region of object constraint and an outer-region of object constraint, over the target cross-attention maps to gradually update the noisy video sample at arbitrary denoising timestep $t$ such that the spatial location of edited objects will be aligned with the target regions.
	
	\textbf{Inner-Region of Object Constraint.}
	To ensure the edited objects approach the user-specified regions, an intuitive objective is to ensure that high responses of cross-attention maps are in the target regions.
	Thus, we can build the inner-region of object constraint $\mathcal{L}_{\text{IR}}(t)$ at denoising timestep $t$:
	\begin{gather}
		\mathcal{L}_{\text{IR}}^{j}(t)=1-\frac{1}{K\times m}\sum_{i=1}^{m} \sum_{k=1}^{K} \text{top}_k(\mathbf{A}_{i}^{j}(t)\times \mathbf{M}_{i}^{j},K),\\ \mathcal{L}_{\text{IR}}(t)=\sum\limits_{j\in \mathcal{I}}\limits \mathcal{L}_{\text{IR}}^{j}(t),
	\end{gather}
	where $\mathcal{L}_{\text{IR}}^{j}(t)$ denotes the constraint corresponding to word index $j\in \mcI$.
	$\mathbf{A}_{i}^{j}(t)$ denotes the a cross-attention map corresponding to word index $j$ in the $i$-th video frame at denoising timestep $t$, where $\mathbf{A}_{i}^{j}(t)\in \mcA^j(t)$ and $\mcA^j(t)=\{\mathbf{A}_1^j(t),\mathbf{A}_2^j(t),\cdots,\mathbf{A}_m^j(t)\}$ is a set of cross-attention maps for word index $j$ in $m$ video frames.
	$\mathbf{M}_{i}^{j}$ denotes the target region mask of the word concept corresponding to word index $j$ in $i$-th video frame.
	$\text{top}_k(\cdot,K)$ represents that $K$ elements with the highest response would be selected, which can reduce the sensitivity of the model to the masks (i.e., no precise masks are required).
	In the experiments, $K$ is set as $20\%$ of the number of the mask regions so that $K$ is adaptively set according to the size of the mask.
	
	\textbf{Outer-Region of Object Constraint.}
	The Inner-region of object constraint can control the edited object to appear inside the mask region, but it cannot ensure that the edited object is not synthesized outside the mask region.
	To mitigate the above issue, we further build a outer-region of object constraint $\mathcal{L}_{\text{OR}}(t)$ at denoising timestep $t$:
	\begin{gather}    \mathcal{L}_{\text{OR}}^{j}(t)\!=\!\frac{1}{K\times m}\sum_{i=1}^{m} \sum_{k=1}^K \text{top}_{k}(\mathbf{A}_{i}^{j}(t)\!\times\! (1\!-\!\mathbf{M}_{i}^{j}),K),\\ \mathcal{L}_{\text{OR}}(t)=\sum\limits_{j\in \mathcal{I}}\limits \mathcal{L}_{\text{OR}}^{j}(t).
	\end{gather}
	Intuitively, $\mathcal{L}_{\text{OR}}(t)$ aims to minimize the activation responses of cross-attention maps out of the mask region, so that $\mathcal{L}_{\text{IR}}(t)$ and $\mathcal{L}_{\text{OR}}(t)$ constrain the cross-attention maps in a complementary manner.

	\textbf{Objective Optimization.}
	We integrate the above constraints to reach the final  RAD objective at denoising timestep $t$:
	$\mcL(t)=\mathcal{L}_{\text{IR}}(t) + \mathcal{L}_{\text{OR}}(t).$
	Then, the noisy video sample $\mcX(t)$ could be updated with a step size of $\alpha_t$ as:
	\begin{equation}
		\mathcal{X}(t)\leftarrow \mathcal{X'}(t)=\mathcal{X}(t)- \alpha_t \nabla\mathcal{L}(t),
	\end{equation}
	where $\alpha_t$ decays linearly at each denoising timestep.
	With the above constraints, $\mcX(t)$ at each timestep gradually moves toward the direction of generating high response attention in the given mask regions, thereby editing the target objects in the user-specified regions.
	
	\subsection{Invariant Region-guided Joint Sampling}\label{Joint Sampling}
\begin{figure}[h]
		\centering{\includegraphics[width=\linewidth]{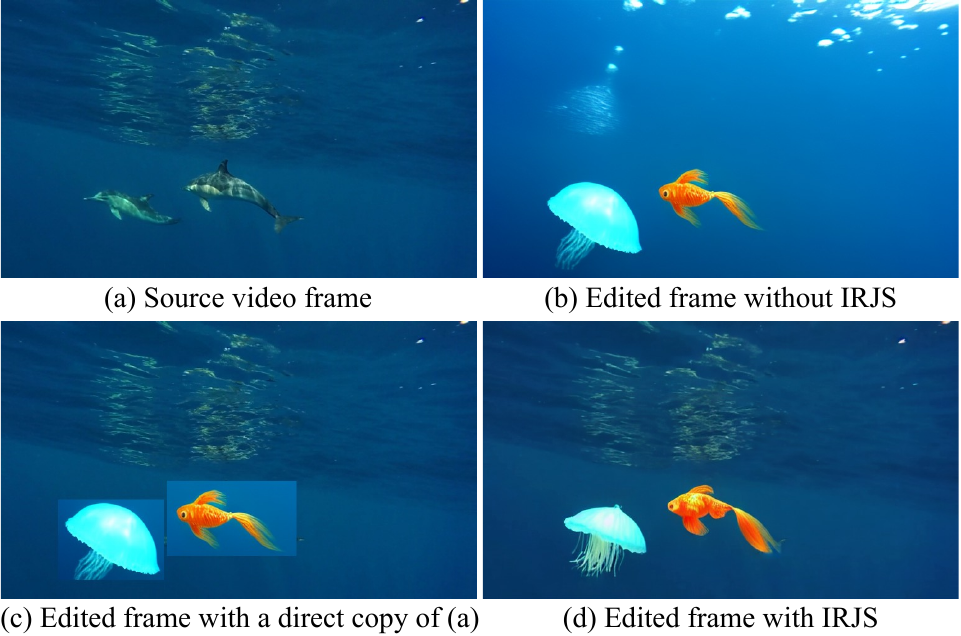}}
		\vspace{-0.6cm}
		\caption{Edited video frames by different methods.
		}
		\label{fig:motivation-BRGJS}
		\vspace{-0.5cm}
	\end{figure}

	The proposed RAD can refocus the cross-attention activation responses to control the editing region. 
	However, we observe that when the user merely wants to edit foreground objects or edit partial foreground objects, e.g., in the case of Fig.~\ref{fig:motivation-BRGJS}, two dolphins need to be edited and the background region is the remaining invariant region, the generated invariant region content is often inconsistent with the original invariant region content.
	As shown in Fig.~\ref{fig:motivation-BRGJS} (b), we can observe that although the edited frame is well-aligned with the edited prompt due to the nice property of RAD, the background region is inconsistent with the one of the source video frame (i.e., Fig.~\ref{fig:motivation-BRGJS} (a)).
	This is because each denoising timestep leads to some sampling errors, and the accumulated errors from all timesteps eventually result in a generated background region that is far from the original background region.
	From the user's perspective, we would like to keep the original invariant background information when manipulating foreground objects.
	To preserve the content of the invariant region during the editing process, a straightforward idea is to copy the corresponding content from the source video directly into the target video, as shown in Fig.~\ref{fig:motivation-BRGJS} (c).
	Intuitively, the object region is not harmonized with the background region, resulting in obvious border artifacts.
	
	\vspace{-0.3cm}
	
	\begin{figure}[h]
		\centering{\includegraphics[width=\linewidth]{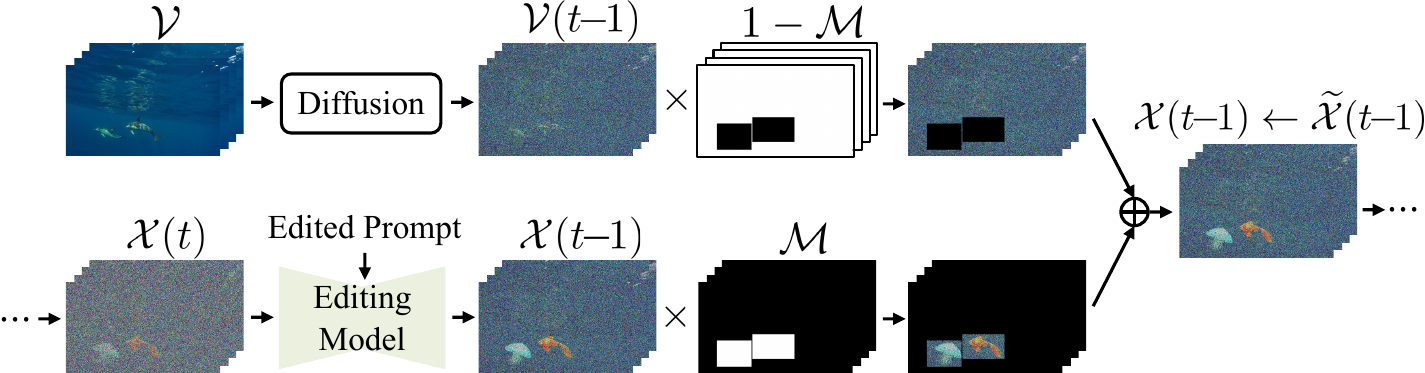}}
		\vspace{-0.6cm}
		\caption{The framework of our proposed IRJS. 
		}
		\vspace{-0.3cm}
		\label{fig:BRGJS}
	\end{figure}
	
	\begin{figure*}[t]
		\centering{\includegraphics[width=\linewidth]{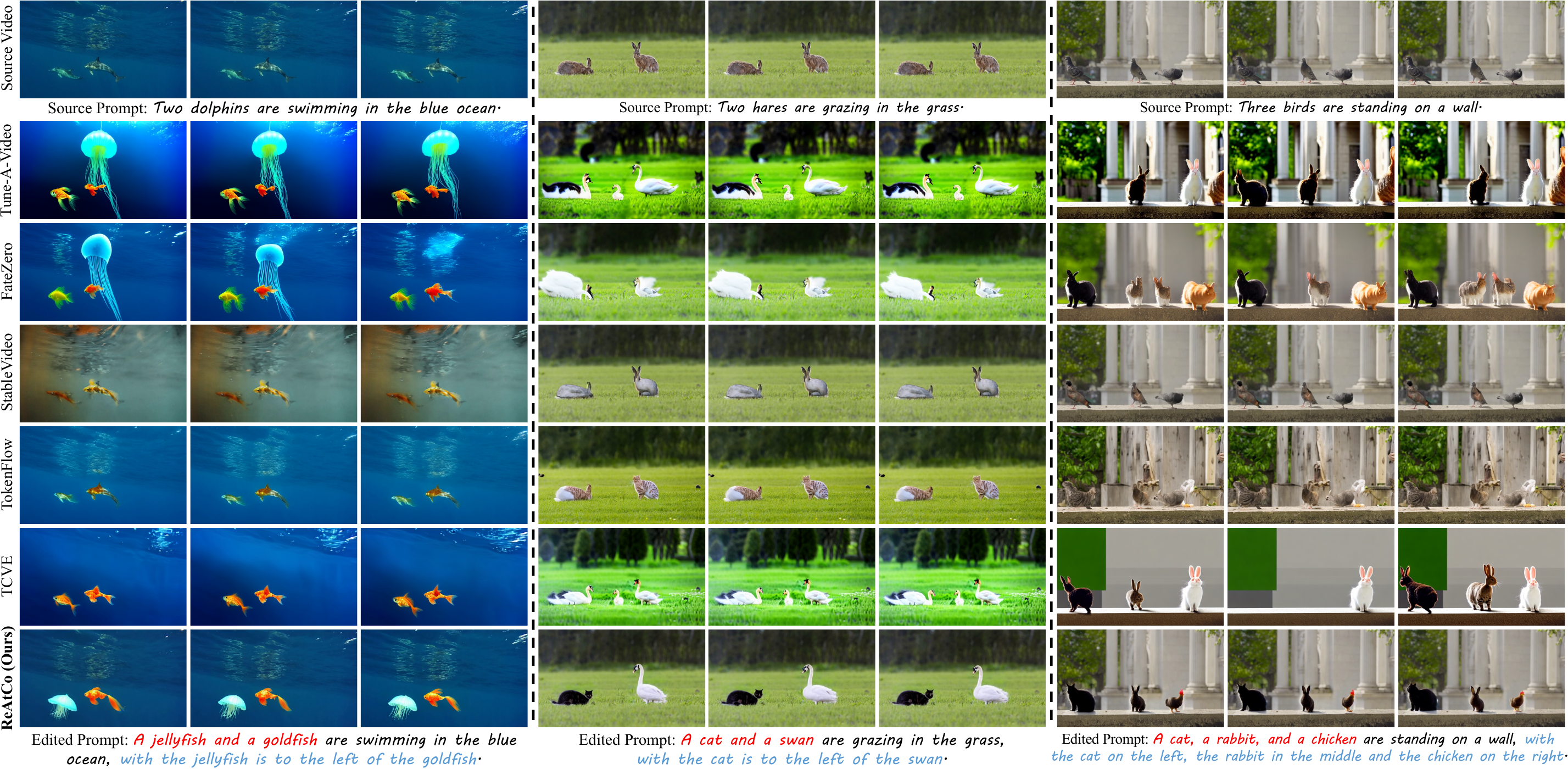}}
		\vspace{-0.6cm}
		\caption{Visual comparisons of different methods in various scenes. Compared with these state-of-the-arts, ReAtCo can edit real-world videos with spatial location alignment, consistent number of objects, and high semantic fidelity.
		}
		\vspace{-0.5cm}
		\label{Qualitative Results}
	\end{figure*}

	To mitigate the above issues, we propose an Invariant Region-guided Joint Sampling (IRJS) strategy to mitigate sampling errors of the invariant region by injecting the original invariant region content into the denoising stage and to constrain the generated content to be harmonized with the original invariant region content.
	The framework of IRJS is illustrated in Fig.~\ref{fig:BRGJS}, where we take the timestep $t$ to $t\!-\!1$ as an example.
	For the vanilla sampling strategy~\cite{ddpm}, the noisy video sample $\mcX(t)$ at timestep $t$ could be denoised into a noisy sample $\mcX(t\!-\!1)$ at timestep $t\!-\!1$ by a video diffusion editing model, but it may disrupt the information of the invariant region.
	The goal of IRJS is to mitigate the sampling error at each timestep by injecting the invariant region of the diffused source video sample into $\mcX(t\!-\!1)$.
	Specifically, the source video $\mcV$ is first diffused into a noisy sample $\mcV(t\!-\!1)$ at timestep $t\!-\!1$ according to predefined diffusion noise scheduler~\cite{ddpm}.
	Then, we use the object masks $\mcM$ (containing all object regions) and invariant region masks $1\!-\!\mcM$ to extract the object region of $\mcX(t\!-\!1)$ and invariant region of $\mcV(t\!-\!1)$, respectively.
	Finally, the extracted regions are added to obtain a noisy sample $\widetilde{\mcX}(t\!-\!1)$:
	\begin{gather}
		\widetilde{\mcX}(t\!-\!1) = \underbrace{\mcX(t\!-\!1)\times \mcM}_{\text{Generated Object Region}} + ~~\underbrace{\mcV(t\!-\!1)\times (1\!-\!\mcM)}_{\text{Original Invariant Region}},
	\end{gather}
	where $\mcX(t\!-\!1)\sim \mcN(\mu_\theta(\mcX(t),t), \Sigma_\theta(\mcX(t),t))$ and $\mcV(t\!-\!1)\sim \mcN(\sqrt{\Bar{\alpha}_t}\mcV(0),(1-\Bar{\alpha}_t)\mathbf{I})$.
	Concretely, $\mu_\theta(\mcX(t),t)$ and $\Sigma_\theta(\mcX(t),t)$ are the predicted parameters of Gaussian transition distribution in the sampling (i.e., denoising) process, and $\Bar{\alpha}_t$ is the total noise variance in the diffusion process predefined by~\cite{ddpm}.
	Further, when the video diffusion editing model is well-trained, then $\mcN(\mu_\theta(\mcX(t),t), \Sigma_\theta(\mcX(t),t))\approx\mcN(\sqrt{\Bar{\alpha}_t}\mcV(0),(1-\Bar{\alpha}_t)\mathbf{I})$.
	This is because the objective of the sampling process is to estimate the transition distribution of the diffusion process at each timestep.
	Thus, we can derive $\widetilde{\mcX}(t\!-\!1)\sim \mcN(\mu_\theta(\mcX(t),t), \Sigma_\theta(\mcX(t),t))$, abided by the distribution of $\mcX(t\!-\!1)$, so that we have $\mcX(t\!-\!1)\leftarrow \widetilde{\mcX}(t\!-\!1)$ that will be used as input for the next iteration in sampling process.

	With multiple iterations of IRJS, the generated object region content could be harmonized with the original invariant region content.
	As shown in Fig.~\ref{fig:motivation-BRGJS} (d), we can observe two benefits: \textbf{1)} the background region (i.e., invariant region) of the edited video is consistent with the source video.
	\textbf{2)} the object region is harmonized with the background region.

	\section{Experiments}\label{experiments}
	
	\subsection{Implementation Details}\label{Implementation Details-main}
	We conduct experiments on the text-guided video editing dataset LOVEU-TGVE-2023~\cite{loveu}, the video samples used in~\cite{Stablevideo}, and the video samples from~\cite{Videvo}.
	Each video has 4 different edited prompts for evaluation.
	For specifying the object regions, we consider enabling the user to provide it in the possibly simplest way, i.e., bounding boxes.
	We consider three standard evaluation metrics that are proposed in the \cite{loveu} to measure the quality of edited videos.
	Frame Consistency is to measure the temporal consistency
	in frames by computing CLIP image embeddings on all frames of output video and reporting the average cosine similarity between all pairs of video frames.
	Textual Alignment is to measure the textual faithfulness of the edited video by computing the average CLIP score between all frames of the output video and the corresponding edited prompt.
	PickScore~\cite{pickscore} is to measure human preference for T2I models.
	We compute the average PickScore in all frames of the output video.
	Furthermore, to measure the spatial location relationships between objects, we introduce the VISOR~\cite{VISOR} that evaluates the spatial relationships (including left, right, above, below) in T2I generation.
	We compute the average VISOR in all frames of the output videos.
	
	\subsection{Baseline Comparisons}
	We compare our ReAtCo with the current state-of-the-arts, including the pioneer in efficient T2I-based video diffusion editing Tune-A-Video~\cite{TAV}, the fusing attention mechanism-based method FateZero~\cite{fatezero}, the atlas model-based method StableVideo~\cite{Stablevideo}, the dual-Unet architecture-based method TCVE~\cite{TCVE}, and the propagation mechanism-based method TokenFlow~\cite{TokenFlow}.
	Below, we analyze quantitative and qualitative experiments.
	
	\textbf{Quantitative results.} Tab.~\ref{tab:1} lists the quantitative results of different methods.
	From these results, we can observe that ReAtCo achieves the best video editing performance under four evaluation metrics.
	In particular, ReAtCo gains considerable performance improvements in the VISOR metric used to measure the spatial location relationships between objects.
	This observation could be ascribed to the fact that ReAtCo can control the spatial location of the edited objects by the well-designed RAD.
	Further analysis of the generated videos is provided in the next part.
	
	\begin{table}[htb]
		\centering
		\caption{Quantitative comparison with evaluated baselines.}\label{tab:1}
		\vspace{-0.3cm}
		\setlength{\tabcolsep}{1.5pt}
		\scalebox{0.8}{
			\begin{tabular}{c|c|c|c|c}
				\hline
				Methods & Frame Consistency & Textual Alignment &  PickScore & VISOR \\
				\hline
				\hline
				Tune-A-Video &92.54  & 26.75   &20.37& 15.62 \\    FateZero &  93.20 &  26.27 & 20.42& 19.37 \\ StableVideo &  93.86 &  24.41 & 19.45 & 10.31\\
				TokenFlow & 94.66 & 26.89 & 20.57 & 11.56\\
				TCVE& 94.79 & 27.71 & 20.58 & 25.31\\
				ReAtCo (Ours)& \textbf{95.24} & \textbf{28.64} & \textbf{20.70} & \textbf{70.62}\\
				\hline
		\end{tabular}}
	\vspace{-0.5cm}
	\end{table}

	\textbf{Qualitative results.}
	We showcase some visual comparison of our ReAtCo against four baselines in Fig.~\ref{Qualitative Results}.
	For the first sample, Tune-A-Video and FateZero suffer from mislocated objects (i.e., the jellyfish is above the goldfish, which is unaligned with \textit{``the jellyfish is to the left of the goldfish''}) and incorrect number of objects (i.e., a jellyfish and two goldfishes are generated, which is inconsistent with \textit{``a jellyfish and a goldfish''}).
	StableVideo, TokenFlow, and TCVE fail to edit the dolphin as the jellyfish.
	In contrast, ReAtCo can output a spatially location-aligned and semantically high-fidelity edited video.
	For the second sample, it is evident that only ReAtCo can faithfully modify the hare on the left to a cat and the hare on the right to a swan.
	For the third sample, this is a more complex scenario containing three foreground objects.
	Therefore, this more challenging case further examines the controllability and robustness of video editing.
	From the results, we can observe that ReAtCo successfully manipulates the bird on the left to a cat, edits the bird in the middle to a rabbit, and modifies the bird on the right to a chicken, while other methods suffer varying degrees of editing failures.
	The above phenomenon is attributed to our proposed RAD.
	At the same time, due to the benefit of IRJS, ReAtCo can also preserve the original background content and there are no obvious border artifacts between the foreground and background regions, showing better harmonization.

	\subsection{Ablation Studies}
	\textbf{Quantitative analysis.}
	We evaluate the effects of the key components in ReAtCo, including RAD and IRJS.
	The results are reported in Tab.~\ref{tab:Ablation}, we conclude the conclusions as: \textbf{1)} Editing videos with RAD is effective, this is because RAD can empower the video diffusion editing model to perceive the spatial location of the foreground objects, thus improving the controllability and performance of video editing.
	Further, IRJS can bring some performance improvement by maintaining information in the invariant region and constraining the generated content to be harmonized with the invariant region.
	\textbf{2)} Combining RAD with IRJS brings further benefits, which proves that editing objects while maintaining invariant region content is feasible and effective.

	\begin{table}[htb]
		\vspace{-0.3cm}
		\centering
		\setlength{\tabcolsep}{1.5pt}
		\caption{Ablation study of the key components in ReAtCo.}\label{tab:Ablation}
		\vspace{-0.3cm}
		\scalebox{0.85}{
			\begin{tabular}{cc|c|c|c|c}
				\hline
				RAD & IRJS &  Frame Consistency  & Textual Alignment  & PickScore & VISOR \\
				\hline
				\hline
				$\checkmark$ & $\checkmark$  & \textbf{95.24} & \textbf{28.64} & \textbf{20.70} & \textbf{70.62}\\
				$\checkmark$ & $\times$  & 94.59 & 28.54 & 20.64& 69.06\\
				$\times$ & $\checkmark$  & 92.97& 26.94&20.45 & 16.25\\
				$\times$ & $\times$  &92.54  & 26.75   &20.37& 15.62\\
				\hline
		\end{tabular}}
	\vspace{-0.3cm}
	\end{table}

\begin{figure}[h]
		\centering{\includegraphics[width=\linewidth]{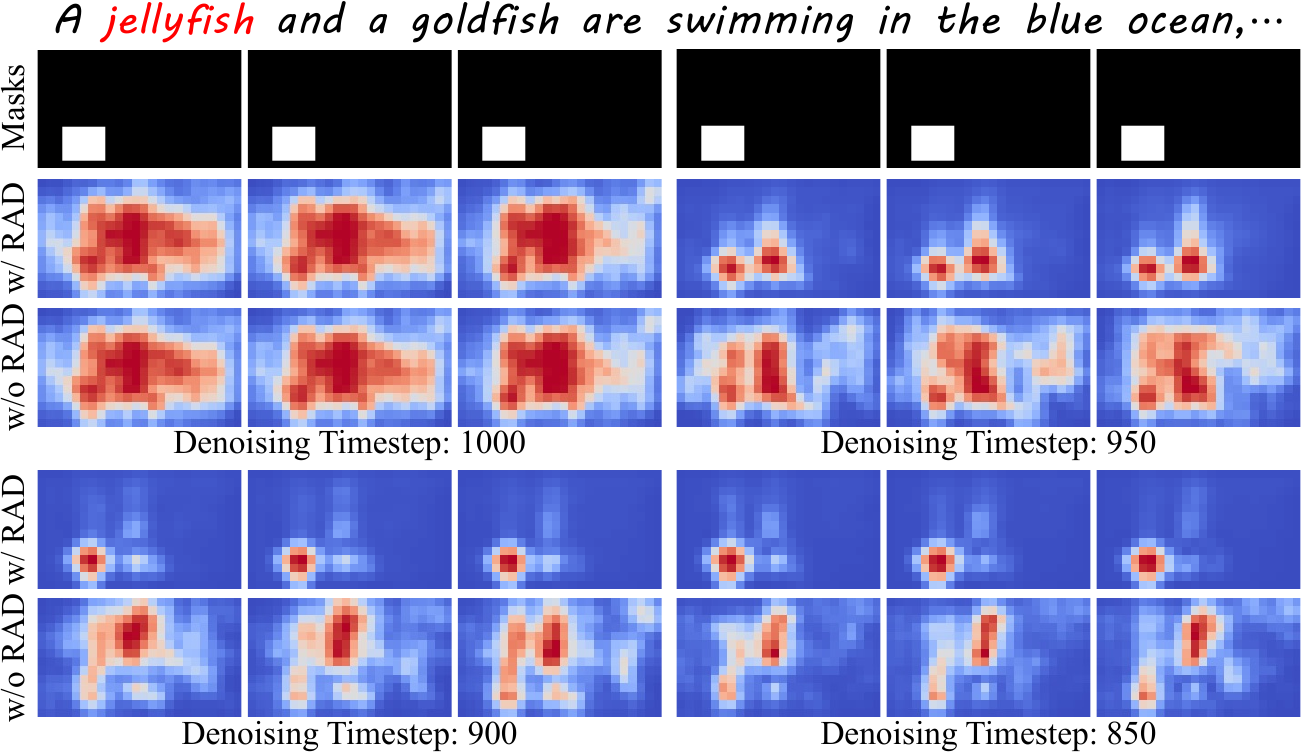}}
		\vspace{-0.5cm}
		\caption{Visualization of cross-attention maps. 
		}
		\vspace{-0.3cm}
		\label{vis_attention}
	\end{figure}

	\textbf{Visualization of cross-attention maps.}
	We take the \textit{``jellyfish''} in the first sample in Fig.~\ref{Qualitative Results} as an example to visualize the cross-attention maps during the denoising process.
	Fig.~\ref{vis_attention} shows the visualization of cross-attention maps associated with the word \textit{``jellyfish''} from \textbf{w/ RAD} and \textbf{w/o RAD}, we can observe that the cross-attention responses in the initial denoising timestep (i.e., denoising timestep is 1000) are all in an irregular state.
	As the denoising timestep decreases, the cross-attention responses gradually focus on a region.
	In particular, for \textbf{w/o RAD}, the focused region of cross-attention responses gradually deviates from the user-specified region of the jellyfish.
	In contrast, cross-attention responses from \textbf{w/ RAD} gradually focus on the user-specified region of the jellyfish, which supports the effectiveness of RAD in refocusing cross-attention activation responses.
	
	\vspace{-0.3cm}
	
\begin{figure}[!htb]
		\centering{\includegraphics[width=\linewidth]{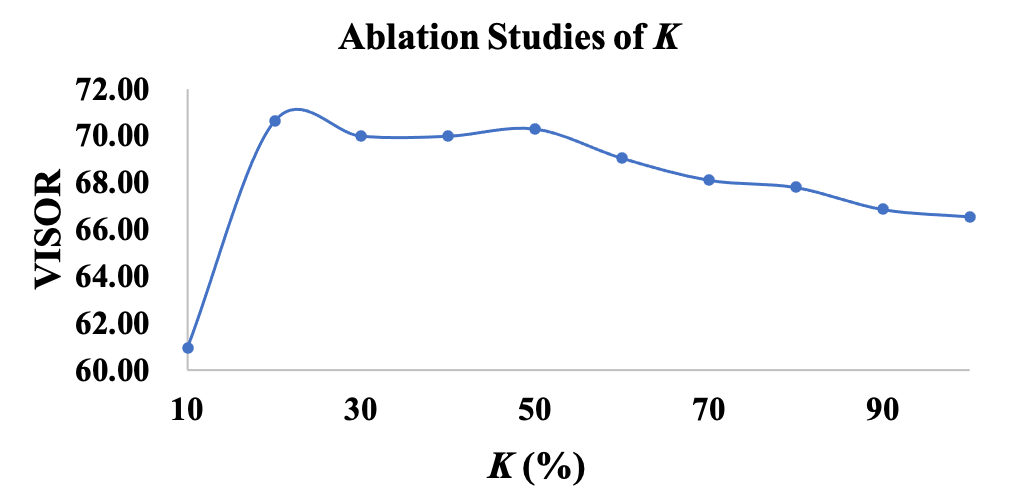}}
		\vspace{-0.9cm}
		\caption{Ablation Studies on various $K$ in $\text{top}_k(\cdot,K)$.
		}
		\label{topk}
		\vspace{-0.5cm}
	\end{figure}

	\textbf{Exploring the effective $K$ in $\text{top}_k(\cdot,K)$.}
	We conduct the ablation studies to explore the effective $K$ in $\text{top}_k(\cdot,K)$.
	Fig.~\ref{topk} illustrates the performance of our method with various $K$ in $\text{top}_k(\cdot,K)$ under VISOR metric, and we can observe that the best performance is reached when $K$ is set to $20\%$.
	Subsequently, the performance is degraded as $K$ increases.
	The above phenomenon demonstrates the fact that the constraints in RAD performed on all responses in the cross-attention maps may lead to a degradation of video editing and the constraints performed on only a few elements with high responses are sufficient to control the region of editing.
	
	\vspace{-0.3cm}

	\section{Conclusion}
	In this paper, we have proposed a Re-Attentional Controllable Video Diffusion Editing (ReAtCo) method for text-guided video editing. 
	ReAtCo is inspired by the observation that the controllability of existing editing methods is not enough, especially in the controllability of spatial location.
	To efficiently improve the controllability of video editing, ReAtCo refocuses the cross-attention activation responses by the well-designed RAD to control the spatial location of the edited objects aligned with the edited text prompts in a training-free manner.
	In particular, we design an IRJS to preserve the invariant region information during editing and to constrain the generated content to be harmonized with the invariant region.
	Extensive experiments demonstrate the effectiveness of our ReAtCo.
	
	\section{Acknowledgement}
	This work was supported by the National Natural Science Foundation of China (Grants No. 62476133), the Research Grants Council of Hong Kong (Collaborative Research Fund No. C7055-21GF) and by the Hong Kong Scholars Program, the Natural Science Foundation of Shandong Province (Grant No. ZR2022LZH003).

	\bibliography{aaai25}
	
	\newpage
	
	\section{Appendix}
	
	\subsection{More Implementation Details}
	
	ReAtCo considers the classic publicly available Tune-A-Video~\cite{TAV} as the pretrained video diffusion editing model that adopts the Stable Diffusion v1.4 as the base model, and the number of denoising steps is fixed as 50.
	$\alpha_t$ decays linearly from 1 to 0.5 during the denoising process.
	We operate the RAD on the cross-attention maps with a resolution of $\frac{H}{32}\times \frac{W}{32}$ ($H$ and $W$ are the height and width of the source video) due to the sufficient semantic information~\cite{p2p}.
	Since the pretrained video diffusion editing model is based on the Latent Diffusion Model paradigm~\cite{stablediffusion}, the proposed RAD and IRJS are performed in the latent space of an autoencoder in practice.
	
	\subsection{Ablation Study of IRJS in the Invariant Region}
	We now evaluate the effects of IRJS in the invariant region to prove the effectiveness of maintaining (i.e., reconstructing) invariant region content.
	To quantify the performance of IRJS for reconstructing invariant region, we consider two evaluation metrics: Peak Signal to Noise Ratio (PSNR) and Learned Perceptual Image Patch Similarity (LPIPS)~\cite{lpips}.
	Tab.~\ref{tab:irjs} reports the PSNR and LPIPS of Ours and Ours w/o IRJS, we can observe that the fidelity of the invariant region shows a severe degradation when the IRJS is removed.
	These results support the fact that our proposed IRJS effectively maintains the invariant region content during video editing. 
	
	\begin{table}[htb]
		\centering
		\caption{Ablation study of IRJS in the invariant region under PSNR and LPIPS metrics.}\label{tab:irjs}
		\scalebox{1.0}{
			\begin{tabular}{c|c|c}
				\hline
				Methods & PSNR (dB)~$\uparrow$ & LPIPS~$\downarrow$  \\
				\hline
				\hline
				Ours w/o IRJS & 29.29   & 0.2706    \\    
				\textbf{Ours} & \textbf{36.57} & \textbf{0.0206} \\
				\hline
		\end{tabular}}
	\end{table}
	
	\subsection{Selection of Words of Interest}\label{Selection Words}
\begin{figure}[!htb]
		\centering{\includegraphics[width=\linewidth]{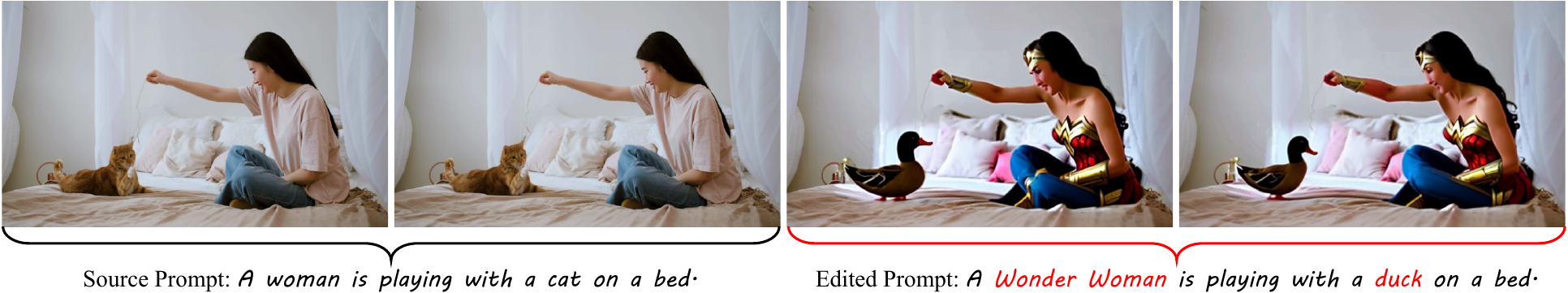}}
		\caption{An example of controlling the objects in the form of compound nouns. We only select \textit{``Woman''} as a word of interest to control the synthesis of Wonder Woman.
		}
		\label{select-words}
	\end{figure}

	Selecting words of interest is an important step in our RAD.
	Typically, given a source prompt and an edited prompt such as \textit{``Two dolphins are swimming in the blue ocean.''} and \textit{``A jellyfish and a goldfish are swimming in the blue ocean.''}, the words of interest could be easily selected as \textit{``jellyfish''} and \textit{``goldfish''}, which is enough to extract the corresponding cross-attention maps for RAD.
	However, in some cases, the user is interested in controlling the objects in the form of compound nouns.
	For example, given a source prompt and an edited prompt, i.e., \textit{``A woman is playing with a cat on a bed.''} and \textit{``A Wonder Woman is playing with a duck on a bed.''}, we can observe that the goal is to change \textit{``woman''} to \textit{``Wonder Woman''} and \textit{``cat''} to \textit{``duck''}.
	A question arises: \textit{how to perform RAD with two cross-attention maps for a single object?}
	In the experiments, we found that a single word almost dominates the target semantic.
	As shown in Fig.~\ref{select-words}, to control the synthesis of Wonder Woman, we only select \textit{``Woman''} as a word of interest, which is enough for RAD to constrain the Wonder Woman within the object region.

	\subsection{Resource-friendly ReAtCo Paradigm}
	In our ReAtCo, we need to track the gradient across the whole big model for attentional control, which inevitably increases GPU memory usage.
	 For example, for a consumer-grade NVIDIA RTX 3090/4090 GPU, only 4 video frames can be processed simultaneously. 
	A naive way is to edit a complete video clip independently every 4 frames, but this straightforward way would definitely disrupt the temporal consistency of the generated video.
	To address this issue, we introduce the long video generation technology~\cite{Gen-L-Video} that can mitigate the temporal inconsistency between multiple generated video clips.
	The basic principle of this technique is to generate multiple short video clips in a sliding-window manner and to expect duplicate video frames at the time nodes, thereby enhancing the temporal consistency between the generated short videos (more details could be found in~\cite{Gen-L-Video} and its publicly available codes). 
	We integrate this technology into our ReAtCo to form a resource-friendly ReAtCo version.
	
	\subsection{Limitations and Discussion}\label{Limitations}
	Limited by the generation capability of the backbone model we used (i.e., Tune-A-Video~\cite{TAV}), some edited results may have some temporal unsmoothness. 
	This limitation could be mitigated by replacing a more powerful backbone model (e.g., if the Sora\footnote{https://openai.com/index/sora/} could be open-sourced) due to the fact that our ReAtCo is a controllable video diffusion editing framework in which the base video model could be used in a plug-and-play manner.
	
	\subsection{Broader Impact}
	The advancement of text-guided video editing will ease the creative efforts of artists and designers, while also causing a risk of misinformation, leading to permanent damage to the reliability of videos. However, it is possible to train a classifier to distinguish the real and ReAtCo-edited videos according to the texture features.

\end{document}